\documentclass[twoside,11pt]{article}
\usepackage[preprint]{dmlr2e}
\usepackage{etoolbox}
\usepackage[T1]{fontenc}
\usepackage[utf8]{inputenc}
\usepackage{lmodern,microtype}

\usepackage{booktabs,tabularx,array,amsmath}
\newcolumntype{Y}{>{\raggedright\arraybackslash}X}

\usepackage{pgfplots}
\usepgfplotslibrary{groupplots}
\usetikzlibrary{arrows.meta,positioning}
\pgfplotsset{compat=1.18}
\definecolor{forgeblue}{HTML}{256788}
\definecolor{forgeorange}{HTML}{A45725}
\definecolor{forgepurple}{HTML}{54416F}
\hypersetup{hidelinks,
  pdftitle={Arya Forge: Auditing Validation and Release Approval for Financial Instruction Data},
  pdfauthor={Kushagra Bhatnagar and Deekshith Marla},
  pdfsubject={Financial dataset validation and release governance},
  pdfkeywords={financial instruction data, dataset validation, provenance, release governance, reproducibility}}
\usepackage{enumitem}
\setlist{nosep}
\newcommand{\artifact}[1]{\texttt{\small\path{#1}}}
\newsavebox{\forgefigurebox}
\newcommand{\fitforgefigure}{%
  \ifdim\wd\forgefigurebox>\linewidth
    \resizebox{\linewidth}{!}{\usebox{\forgefigurebox}}%
  \else\usebox{\forgefigurebox}\fi}
\AtBeginEnvironment{figure}{\setlength{\abovecaptionskip}{9pt}\setlength{\belowcaptionskip}{2pt}}
\title{Arya Forge: Auditing Validation and Release Approval for Financial Instruction Data}
\author{Kushagra Bhatnagar\hspace{24pt}Deekshith Marla}
\firstpageno{1}
\definecolor{aryateal}{HTML}{286667}
\definecolor{aryagray}{HTML}{5C6263}
\graphicspath{{./}{manuscript/}}
\makeatletter
\def\ps@arya{%
  \def\@oddhead{\normalfont\fontsize{9}{11}\selectfont\color{aryagray}%
    Arya.ai\hfil Arya Forge}%
  \let\@evenhead\@oddhead
  \def\@oddfoot{\normalfont\small\hfil\thepage\hfil}%
  \let\@evenfoot\@oddfoot}
\renewcommand{\maketitle}{%
  \thispagestyle{plain}%
  \vspace*{-33pt}%
  \noindent\includegraphics[width=80pt]{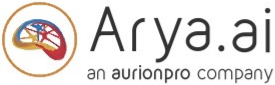}%
  \hfill{\color{aryagray}\fontsize{9}{11}\selectfont\@date}\par
  \vspace{9pt}%
  {\color{black!18}\hrule height .45pt}\vspace{14pt}%
  {\raggedright\bfseries\fontsize{30}{34}\selectfont Arya Forge:\par}%
  \vspace{5pt}%
  {\raggedright\bfseries\fontsize{22}{28}\selectfont
   Auditing Validation and Release Approval\\
   for Financial Instruction Data\par}%
  \vspace{17pt}%
  {\noindent\bfseries\fontsize{12.5}{15}\selectfont\@author\par}%
  \vspace{6pt}%
  {\noindent\fontsize{11}{14}\selectfont Arya.ai\par}%
  \vspace{19pt}%
}
\renewenvironment{abstract}{%
  \par\noindent{\color{aryateal}\bfseries\normalsize ABSTRACT}\par
  \vspace{6pt}\noindent\ignorespaces
}{\par\vspace{4pt}}
\renewenvironment{keywords}{%
  \par\noindent\small\textbf{Keywords: }\ignorespaces
}{\par\vspace{8pt}}
\renewcommand{\section}{\@startsiction{section}{1}{\z@}%
  {-2.5ex plus -1ex minus -.2ex}{1.2ex plus .2ex}%
  {\normalfont\large\bfseries\color{aryateal}}}
\renewcommand{\subsection}{\@startsection{subsection}{2}{\z@}%
  {-2ex plus -.5ex minus -.2ex}{.9ex plus .2ex}%
  {\normalfont\normalsize\bfseries}}
\makeatother
\date{25 October 2026}
\begin{document}
\raggedbottom
\maketitle

\begin{abstract}
Financial instruction-data pipelines must distinguish content validity from permission to release a particular dataset. Arya Forge combines typed records, current-content checks, and approval coverage bound to a dataset manifest. We evaluate its release boundaries in a financial-data systems case study. An integrity audit covers 100,000 synthetic instruction records; a separate arithmetic implementation agrees with all 15,000 numeric records across twelve formula families. In 240 scenarios with 2,400 treatment rows, adding financial and release-record checks to a short-instruction filtering recipe rejects all injected numeric faults and policy holds while retaining 220 of 240 clean controls. The twenty lost controls comprise every tested cash-conversion-cycle example. Nineteen persisted-release cases show that eleven selected negative cases are blocked, four equivalent representations are rejected, and changes to external evidence after binding are outside ordinary Gold promotion's checks. Local processing measurements and controlled checkpoint resumptions characterize cost and recovery through 100,000 input rows. The results expose a tradeoff between strict producer-contract enforcement and useful-representation retention, and identify the protected inputs required for manifest-bound approval. These findings apply to the tested implementation and workloads; they do not establish general financial correctness, authenticated approval, expert-validated quality, or downstream model improvement.
\end{abstract}

\section{Introduction}
Financial instruction data supply both answers and examples of reasoning to language models. A record can pass a schema check while its answer has changed since validation, its cited evidence applies to another period, or its approval covers different content. Conversely, a valid explanation can fail a strict producer contract because it expresses the same quantity differently. Release controls therefore need to be evaluated for both invalid acceptance and useful-content loss.

Consider a debt-service coverage ratio example whose stored calculation flag remains PASS after the assistant answer is edited. A release decision based only on that flag can admit stale supervision. Recomputing supported quantities and checking the released response can reject the mutation. Yet requiring the factory's exact wording also rejects some valid alternatives. Binding approval to the resulting manifest addresses a separate question: whether the recorded decision covers this particular release. Arithmetic, representation compatibility, and authorization are distinct properties.

Arya Forge implements auditing, synthetic instruction generation, qualification, deduplication, selection, and dataset promotion. We study three questions: (RQ1) which checks and trusted inputs connect financial-record validation and approval coverage to the released artifact; (RQ2) which constructed failures are rejected and which valid representations are lost; and (RQ3) what local processing cost and checkpoint-recovery behavior are observed? Independent financial quality and downstream model utility remain separate evaluation targets.

\begin{samepage}
The paper makes three contributions:
\begin{enumerate}
\item A specification of the evaluated release interfaces, distinguishing fresh content checks, stored validation assertions, unresolved findings, and manifest-scoped approval coverage.
\item An empirical analysis of enforcement and retention: component diagnostics, paired filter treatments, and persisted-release cases expose both blocked faults and rejected valid representations.
\item Supporting integrity, processing, and recovery measurements, with an explicit distinction between portable code checks and privately retained experimental evidence.
\end{enumerate}
\end{samepage}
The contribution is the measured interaction of these mechanisms in financial instruction-data production. Artifact binding, executable validation, and release gates are established techniques. We do not claim a new cryptographic primitive or infer a model benefit from results elsewhere in the larger repository. An exploratory AI-assisted record review is reported separately in Appendix~\ref{app:ai}.

\section{Related work}
\paragraph{Data processing and validation.} Data-Juicer 2.0 and DataTrove provide extensible processing infrastructure \citep{datajuicer,datatrove}; both can host application-specific checks. TensorFlow Data Validation and Deequ establish schema, anomaly, and executable constraint checking \citep{tfdv,deequ}. Our filter experiment measures two pinned DataTrove blocks with added Forge mechanisms. It does not compare equally informed financial validators or rank processing frameworks. Dolma and datasheets motivate explicit curation and artifact documentation \citep{dolma,datasheets}. DataComp-LM evaluates data choices under controlled training recipes \citep{dclm}; comparable downstream evidence is outside this study.

\paragraph{Evidence and release control.} in-toto binds authorized supply-chain steps to artifacts through signed metadata \citep{intoto}. Attesting LLM Pipelines proposes an attestation-aware admission gate and an evaluation blueprint for training and release claims \citep{attesting}. Reusing Assurance Evidence formalizes candidate-relative evidence validity, implements signed checks, and studies stale dependency reuse in public DVC pipelines \citep{reuse}. Structure Before Intelligence evaluates how validation, recovery, and release policies affect invalid outputs, yield, and cost in scientific workflows \citep{structure}. These works establish release gating and evidence binding as prior art. Forge examines financial instruction-record contracts and approval coverage, including their compatibility costs; it provides neither signature authentication nor incremental dependency analysis. Table~\ref{tab:related} locates the empirical distinction. The two October 2026 papers were incorporated during manuscript revision after our original experiments.

The Data Provenance Initiative audits dataset licensing and attribution \citep{provenance}; an approval assertion in Forge does not independently establish either. OriginBlame propagates record- and token-level provenance and resolves revocation queries \citep{originblame}. Forge instead evaluates eligibility at a particular dataset-release boundary, with no external revocation discovery service.

\paragraph{Financial instruction data.} FinQA supplies expert-authored questions and reasoning programs; FinanceBench supplies evidence-grounded financial QA \citep{finqa,financebench}. PIXIU and FinDAP study financial instruction data, capabilities, and model adaptation \citep{pixiu,findap}. From Documents to Reasoning combines synthetic financial QA validation with downstream adaptation and compares evaluated arithmetic expressions to reduce sensitivity to answer representation \citep{documents}. Forge's exact native text projection enforces a narrower producer contract and rejects some equivalent wording. Financial generation and arithmetic checking alone are therefore insufficient novelty claims. External QA datasets would require compatible adapters, documented source-overlap checks, and explicit accounting of unsupported cases.

\begin{table}[!htb]
\centering\small
\caption{Closest release-control comparisons. Entries describe the cited studies, not exhaustive limits of their software.}
\label{tab:related}
\begin{tabularx}{\textwidth}{>{\raggedright\arraybackslash}p{3cm}YY}
\toprule Study & Evidence and decision unit & Relation to the present evaluation\\ \midrule
in-toto & Signed supply-chain steps and artifact identity. & Establishes authenticated step binding; Forge trusts supplied authority assertions.\\
Attesting LLM Pipelines & Attested training/release claims at admission. & Gate design and evaluation blueprint; Forge reports executed record and release probes.\\
Reusing Assurance Evidence & Signed control evidence valid for a candidate and its dependencies. & Studies stale evidence reuse; Forge tests current financial records and approval scope.\\
Structure Before Intelligence & Validation and release of workflow outputs. & Measures safety, yield, and cost; Forge examines instruction-data representation and approval coverage.\\
Arya Forge & Typed instruction records and approval bound to a manifest. & Measures selected faults, clean-template losses, and equivalent-representation rejection; no authority authentication.\\ \bottomrule
\end{tabularx}
\end{table}

\section{System and trust boundary}
\subsection{Typed records and stages}
\begin{figure}[!htb]
\centering
\begin{lrbox}{\forgefigurebox}
\begin{tikzpicture}[
  stage/.style={draw=forgeblue,rounded corners=2pt,fill=forgeblue!7,align=center,text width=2.45cm,minimum height=1.05cm,font=\small},
  evidence/.style={draw=forgeorange,rounded corners=2pt,fill=forgeorange!7,align=center,text width=3.15cm,minimum height=1.05cm,font=\small},
  flow/.style={-{Latex[length=2mm]},line width=.65pt},
  note/.style={font=\scriptsize,align=center,fill=white,inner sep=2pt}]
\node[stage] (raw) at (0,0) {Raw\\Source bytes};
\node[stage] (bronze) at (3.5,0) {Bronze\\Typed records};
\node[stage] (silver) at (7,0) {Silver\\Candidate release};
\node[stage] (gold) at (10.5,0) {Gold\\Approved release};
\draw[flow] (raw) -- node[note,above=7mm] {migrate} (bronze);
\draw[flow] (bronze) -- node[note,above=7mm] {qualify/select} (silver);
\draw[flow] (silver) -- node[note,above=7mm] {check/promote} (gold);
\node[evidence] (evidence) at (3.5,-2.7) {External evidence\\Authority assertions};
\node[evidence] (registry) at (10.5,-2.7) {Bound registry\\Unique active coverage};
\draw[flow] (evidence) -- node[note,above=1mm,text width=3.2cm] {bind: verify evidence bytes\\and pin Silver manifest} (registry);
\draw[flow,dashed] (registry) -- node[note,right] {trusted\\registry} (gold);
\draw[flow,dashed] (silver.south) -- node[note,pos=.45,right] {manifest identity} (registry.north west);
\end{tikzpicture}
\end{lrbox}
\fitforgefigure
\caption{Dataset stages and the ordinary approval path. Solid arrows denote processing; dashed arrows denote inputs to binding or promotion. Ordinary Gold promotion rechecks current records, parent integrity, and active coverage, but does not reopen external evidence. Supplemental promotion has a different evidence-read boundary (Section~\ref{sec:coverage}). Authority assertions, accepted identities, and evaluation time remain trusted inputs.}
\label{fig:architecture}
\end{figure}
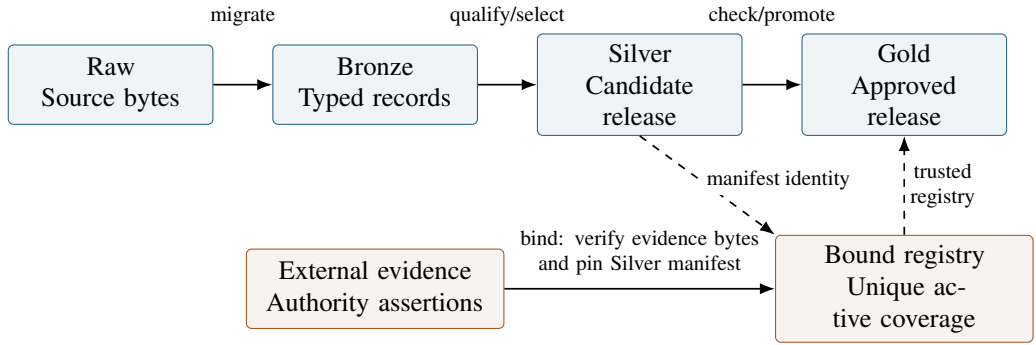

Forge defines strict record contracts for continued pretraining and supervised fine-tuning. They carry content, task metadata, jurisdiction, temporal fields, provenance, rights assertions, quality and validation metadata, and transformation lineage. Figure~\ref{fig:architecture} shows the stages and the approval boundary.
Raw denotes source bytes and inventory; Bronze denotes canonical migration; Silver denotes selected and validated candidate content; Gold denotes a release with explicit approval coverage. The protocol intends Gold to be an approval state, not a synonym for high quality. Historical implementation gaps described below show why that intention must be tested at every promotion entry point.

Releases include JSONL and Parquet representations, statistics, source inventories, rights and quality reports, and manifests. Manifests bind artifact hashes, record-set hashes, configuration, validator versions, and parent releases. Output construction uses a staging directory and refuses an existing immutable destination. These controls detect divergence relative to the recorded hashes; they do not prove that the source facts or approvals were correct when recorded.

\subsection{Approval coverage}\label{sec:coverage}
The promotion registry binds to an exact Silver manifest. Each retained record must be covered by one active source- or record-scoped decision, with no missing or ambiguous coverage. Decision metadata includes an approval reference, evidence digest, permission assertions, obligations, and validity information. A named governance authority remains responsible for the authenticity and substance of the decision.

Let $r$ be a record, $m$ its parent manifest, $g$ the supplied approval registry, and $t$ the evaluation time. Write $C_v(r)$ for the supported current-content checks under validator version $v$, $Q(r)$ for absence of unresolved quarantine conditions, $A(r,m,g,t)$ for unique active approval coverage, and $I(m)$ for manifest-listed artifact integrity. The intended promotion condition is
\[
\operatorname{Promote}(r,m,g,t) \Rightarrow C_v(r) \land Q(r) \land A(r,m,g,t) \land I(m).
\]
This is a design requirement, not a formal proof about all program paths. Approval resolves only the rights conditions within its scope. It must not resolve an unrelated privacy, arithmetic, or source-applicability finding. Table~\ref{tab:boundary} specifies the tested transition sequence.

\begin{table}[!htb]
\centering\small
\caption{Checks in the ordinary governed promotion path. Registry binding and Gold promotion are separate operations.}
\label{tab:boundary}
\begin{tabularx}{\textwidth}{lX}
\toprule Operation & Required evidence and effect\\ \midrule
Load Silver & Check stage and manifest-listed artifact hashes; parse typed records and check count.\\
Bind registry & Check external evidence bytes and decision activity; bind decisions to this manifest and require unique coverage.\\
Evaluate Gold & Rerun current record checks and quarantine; evaluate manifest pin and unique active approval coverage at the supplied time.\\
Write Gold & Record approval and parent lineage, embed registry and coverage report, and refuse an existing immutable destination.\\ \bottomrule
\end{tabularx}
\end{table}

The trusted inputs are the verifier implementation, caller-supplied evaluation time, governance authority, and accepted manifest/registry identities. Activity includes approval/effective dates, exclusive expiry, and a recorded revocation date. The library does not discover external revocations. Hashes detect divergence from an accepted reference; an actor able to replace content and references can create a self-consistent release. The ordinary promotion function accepts a registry file, without a separate expected-registry-digest argument. Protecting that accepted registry and the evaluation time is a caller/deployment responsibility, not an additional authentication check performed by the function.

\paragraph{Evidence freshness.} The ordinary path has a binding-time evidence boundary: binding checks external evidence bytes, and promotion subsequently trusts the bound registry. A deployment using snapshot approval must retain the approved evidence version and protect its reference; the ordinary Gold artifact embeds the registry and coverage report, not the external evidence itself. A deployment requiring current evidence or live revocation status needs an additional freshness mechanism. These are different operational requirements, and neither is established by hashing alone.

\paragraph{Entry-point coverage.} The nineteen persisted-release cases exercise ordinary promotion. Supplemental Gold promotion reuses the current record gate but also reopens referenced corpus approval evidence while deriving its registry, and checks selection/governance lineage. It is covered by software regressions, not by the nineteen-case experiment. The low-level \texttt{DatasetReleaseBuilder} is a trusted writer and does not implement the full approval gate. Consequently, ordinary promotion's accepted post-binding mutation must not be generalized to the supplemental path.

\subsection{What is revalidated at promotion}
Every record entering the shared gate is parsed against the current typed schema. The gate checks score, route, lineage, unresolved recorded findings, and current text against implemented sensitive-data patterns. For recognized native numeric contracts, it recomputes supported quantities and compares the question, derivation, and response with the producer's canonical projection. Unsupported recognized numeric contracts are refused. Unmarked arithmetic elsewhere is not comprehensively discovered.

Task validation also consults stored assertions: schema/task error flags, a numeric-valid flag when calculation is required, and grounding/citation flags when retrieval is required. The shared promotion gate does not rerun every domain-specific generator or qualification validator. Table~\ref{tab:mixture} makes that coverage explicit across the ten audited domains. Fresh privacy-pattern checks and numeric-contract checks should therefore not be read as fresh expert validation of every financial statement.

\subsection{Corpus processing and selection}
The streaming corpus path uses disk-backed fingerprints, bounded audit shards, checkpoint state, and source/configuration pins. Qualification generates rule-level findings and an inventory for rights review. Candidate selection combines eligibility, near-duplicate review, tokenizer accounting, and curriculum limits. The ordinary small-release builder materializes records in memory; the existence of a streaming audit does not imply that every command is memory bounded.

Stored quality scores and deterministic PASS labels are operational signals. Some synthetic factories assign fixed scores rather than estimating calibrated accuracy. In general, absence of a finding means no implemented check rejected the record; it is not proof of factual truth. These distinctions govern all interpretations in this paper.

\section{Evaluation design}
\subsection{Frozen local assets}
The asset audit enumerates ten local Gold releases and verifies manifest-listed artifact digests, JSONL counts, Parquet metadata counts, record-set hashes, and source-manifest hashes. It inventories declared domains and source identifiers. Synthetic status is reported from the explicit source prefixes; it does not independently establish ownership. A subsequent independently written consistency checker decodes every Parquet record payload and compares it with the corresponding JSONL row, rejecting duplicate JSON keys and nonfinite values. It separately reconstructs all twelve projected Parquet columns from JSONL and checks their types and values; this supplements the original count-and-hash audit.

The audit does not rewrite historical data or replace their original validation metadata. A new report binds the current verifier source hashes and runtime versions to its observations. Because the working tree contains unrelated ongoing changes, the research package records the base revision and exact relevant source snapshots; a bare Git commit alone is not claimed to reproduce the repaired state.

\subsection{Separate arithmetic implementation}
The historical numeric factory set \texttt{independently\_verified} by serializing a case and repeating the same computation. A deterministic bug can therefore satisfy the old check. The revised default uses a separate module that does not call the production calculator or generator. Ratio, loss, and conversion expressions use rational arithmetic; EMI is recomputed from the sum of discounted instalments; CAGR uses logarithm and exponential evaluation. For principal $P$, monthly rate $i$, and $n$ instalments, the reference computes
\[
\operatorname{EMI}_{\mathrm{ref}} =
\frac{P}{\sum_{t=1}^{n}(1+i)^{-t}}.
\]
At zero interest this reduces to $P/n$. Typed input constraints are enforced before the legacy EMI reference path, including a bounded tenure.

The reference checks output names and units, exact half-up displayed rounding, and unrounded agreement within
\[
\max\left(10^{-20},\; |v_{\mathrm{ref}}|\,10^{-24}\right).
\]
The tolerance accommodates the production implementation's decimal precision. New generation records a new verification version. Explicit legacy reproduction remains available for historical artifacts and retains the old version identifier; those artifacts receive a fresh check before new promotion.

Separate implementation reduces shared-code failure modes, but the implementations retain common mathematical specifications and typed assumptions. Agreement is neither expert validation of those assumptions nor evidence that generated scenarios represent real borrowers.

\subsection{Current-content promotion checks}
New numeric release validation recomputes supported native factory cases, checks their structured results and derivations, and binds the canonical question and assistant response to those results. Changing domain metadata alone cannot disable checks when a native numeric contract or source marker remains. Records recognized by the numeric-contract dispatch are rejected when their format is unsupported; unmarked arithmetic elsewhere is not comprehensively discovered. The native contract is intentionally strict: alternate valid prose may require explicit migration rather than passing automatically.

New quarantine validation scans current training content with the existing sensitive-data patterns and rejects unresolved recorded sensitive-data findings. A generic human-review boolean cannot dismiss a finding. Rights-only quarantine remains compatible with the approval path; unknown or unrelated quarantine reasons remain blocked. The scanner is limited to its implemented patterns and is not a comprehensive privacy detector.

\subsection{Overlap detection and near deduplication}
The revised overlap audit aligns recognized question, request, structured-input, and conversation envelopes. It retains substantive context, numeric values, and prior conversation turns while ignoring absent optional message metadata. It preserves existing identity/source/content checks and rejects empty or missing training inputs rather than returning a vacuous clean result. This is exact normalized overlap detection, not semantic decontamination or a proof of template independence.

Near deduplication now preserves Unicode word characters and financially significant literals. A near-match may reject a record only when its ordered financial-literal signature agrees with that of the candidate. This prevents lexical similarity alone from erasing changed values or signs. Empty token streams cannot near-match. The policy is deliberately conservative: equivalent numeric formatting can retain extra examples, and matching literals do not establish semantic equivalence. The SQLite index stores the normalization and detector policy and rejects incompatible resume state.

\subsection{Targeted component diagnostics}
The diagnostic suite preserves pre-repair tracked source from the recorded repository revision and loads those component snapshots under separate module names. It compares them with the repaired components on author-selected clean controls and fault probes. Cases include deliberately incorrect arithmetic, changed responses with stale flags, unresolved sensitive findings, repeated prompts across formats, and financially distinct near-duplicate candidates.

The cases were authored in response to observed defects. They are retrospective regression diagnostics, not a randomly sampled test set, an independent benchmark, a production incidence estimate, or a comparison with DataTrove/Data-Juicer. The harness reruns the substantive cases to check deterministic equality and records each expectation and observation. The baseline is component replay with shared current dependencies, not a complete historical environment reconstruction.

\subsection{Controlled filter composition}
We freeze a second study before drawing its new parameters and observing filter outcomes. The study remains internally designed after the initial defect analysis, not externally preregistered. It contains 240 test scenarios, twenty per numeric formula, allocated evenly between two supported native answer formats. Each scenario receives ten treatments: one clean control; four numeric inconsistencies affecting response value, response unit, structured truth, or prompt value; four release-policy holds affecting current sensitive content, a retained sensitive finding, a rejected route, or a below-threshold quality score; and one repetition probe. Thus there are 2,400 paired test rows, not 2,400 independent financial situations. An additional 24 development scenarios yield 240 rows. All content is locally generated synthetic material covered by the recorded project-owner authorization; policy holds are not characterized as legally unauthorized source data.

The baseline executes Gopher repetition and quality heuristics \citep{gopher} through the actual \texttt{GopherRepetitionFilter} and \texttt{GopherQualityFilter} blocks from DataTrove 0.10.1. It sees the concatenated system, user, and assistant messages, without hidden quality metadata or expected labels. We report its defaults as a sensitivity analysis and prespecify a short-instruction recipe: minimum five words, minimum average word length one, minimum zero stop words, and \texttt{max\_non\_alpha\_words\_ratio}=0.3; other settings remain at package defaults. The latter implementation parameter acts as a minimum fraction of words containing an alphabetic character. Two further arms append either the numeric validator or the actual Forge release-record gate with affirmative rights eligibility. The final arm includes numeric checks, current and retained sensitive-data findings, route, score, and lineage checks. It does not execute the complete raw-to-Gold pipeline, approval-registry coverage, corpus selection, or either framework's distributed executor.

Test/development numeric parameter hashes are disjoint, and test hashes are compared with the existing numeric Gold tranche. Formula definitions, generators, and answer templates remain shared. Outcomes are clean-control retention and acceptance of each constructed fault/hold category. We retain per-scenario decisions and report three repeated passes; repeated treatments and shared templates preclude treating aggregate counts as independent binomial observations. The experiment does not estimate naturally occurring defect prevalence or valid-paraphrase rejection. Four treatments alter only metadata unavailable to the text baseline: structured truth, a retained sensitive finding, rejected route, and quality score. Baseline acceptance of these rows is expected from its input boundary; it is not evidence that a text filter fails at its intended task.

\subsection{Persisted release-boundary experiment}
A further code-informed protocol freezes nineteen cases and hashes of 100 source and configuration files before runner execution. Every case starts with the same two deterministic native numeric records (DSCR seed 17 and EMI seed 29), builds a persisted Bronze release, and invokes the production Bronze-to-Silver promotion. It then binds explicitly fictional approval evidence and attempts governed Silver-to-Gold promotion. Two cases terminate during evidence binding; seventeen reach a Gold attempt. Dates and identifiers are fixed. These dependent cases test selected mechanisms, not a representative sample or real approval authority.

Cases cover source- and record-scoped controls; incomplete, overlapping, expired, and wrong-parent approvals; dataset-byte tampering; incorrect current answers with stale flags; current and stored sensitive findings; equivalent representations; and evidence changes before or after binding. To isolate content revalidation, content-mutation cases coherently update both dataset formats, artifact and record-set hashes, and bind fresh fictional approval to the changed Silver manifest. This is not a bypass of an unchanged prior approval. Four constrained semantic variants change a prompt synonym, add an answer introduction, spell the ratio unit as ``times,'' or add redundant derivation parentheses; their values and parameters remain fixed. A separate JSON whitespace/member-order control preserves the parsed object.

Accepted outputs are checked for artifact and row consistency, coverage, lineage, training-eligibility metadata, and immutable-destination refusal. All Gold attempts check that their Silver parent remains unchanged. Two complete runs compare result bytes and all persisted file hashes. An initial runner assertion expected lowercase \texttt{sft} instead of the schema's \texttt{SFT}; the instrumentation-only repair and aborted artifacts are retained. No case expectation, protocol, or core implementation changed.

\subsection{Local processing and recovery experiment}
We freeze a separate workload and measurement specification before timing three actual corpus stages: streaming audit, Bronze migration, and route/truth qualification. Each ten-row block contains seven distinct single-chunk CPT records, two distinct SFT records, and one exact duplicate of the second SFT record. The raw workload therefore has 10\% duplicates and a 70:30 CPT:SFT mixture. We run 1,000, 10,000, and 100,000 raw rows with three repetitions per size. This authored fixed-length workload tests execution behavior, not financial correctness or realistic document diversity.

Each stage uses a fresh Python subprocess on an Apple M3 Pro with 18\,GiB RAM, Python 3.13.7, and macOS. Stage time excludes interpreter startup/imports; peak resident memory includes them and is measured using the child's platform-corrected \texttt{ru\_maxrss}. Logical output size includes stage artifacts and checkpoint state. The OS cache is not flushed, and unrelated desktop workloads are uncontrolled. Qualification receives 90\% of raw rows after duplicate exclusion. At the 10K scale, public interruption hooks stop each stage after five committed chunks or shards; a new process resumes it. We compare substantive row-multiset hashes and summaries with the uninterrupted run, allowing provenance paths to differ. Bronze and qualification reuse uninterrupted parent reports, isolating stage-local recovery rather than a chained end-to-end recovery. This tests controlled post-checkpoint recovery, not arbitrary process kills or power loss.

\section{Results}
\subsection{Asset integrity and composition}
The audit verifies all ten releases, all 80 manifest-listed artifacts, and 100,000 instruction records. The subsequent exhaustive format check finds 100,000 matching decoded row payloads and 1,200,000 matching projected field values, with no discrepancies. This establishes representation consistency for these bytes, not factual validity. Table~\ref{tab:mixture} reports the observed mixture. Every record has a synthetic source prefix; none has its stored human-review flag set. The latter describes metadata, not a finding that no person ever reviewed any related material.

\begin{table}[!htb]
\centering
\caption{Audited synthetic Gold mixture and shared promotion-gate coverage. All domains receive current schema, privacy-pattern, and unresolved-finding checks. ``None'' means no additional domain-semantic recomputation in this shared gate; domain-specific generation or qualification checks may have run earlier, and stored validation assertions are still consulted.}
\label{tab:mixture}
\begin{tabularx}{\textwidth}{lrY}
\toprule Domain & Records & Additional fresh semantic check\\ \midrule
Corporate credit & 20,000 & None\\
Numeric reasoning & 15,000 & Supported native arithmetic and canonical text; supported legacy EMI\\
Grounded regulation & 15,000 & None; grounding/citation flags are stored assertions\\
Financial analysis & 15,000 & None\\
SME credit & 10,000 & None\\
Trade finance & 8,000 & None\\
AML/KYC & 7,000 & None\\
Transaction banking & 5,000 & None\\
Refusal/safety & 3,000 & None\\
General retail banking & 2,000 & None\\ \midrule
Total & 100,000 & Arithmetic cross-check covers the 15,000-record numeric tranche\\ \bottomrule
\end{tabularx}
\end{table}

All 15,000 numeric records pass both the separate arithmetic implementation and fresh native-contract validation. There are 1,250 records in each of twelve formula families. The families are EMI, debt-service and interest coverage, fixed-obligation-to-income, loan-to-value, debt-to-equity, net-debt-to-EBITDA, liquidity ratios, cash conversion cycle, expected loss, compound annual growth, and foreign-exchange conversion. Their historical verifier identifier is \texttt{financial-computation-v1}; the new result is a retrospective check of those immutable records, not a reclassification of the original verifier as independent. Exhaustive agreement on this fixed population does not yield an external error rate for other cases, domains, or specifications.

\subsection{Repaired boundaries and exact overlap}
All fifteen targeted diagnostic cases satisfy their expected behavior under the repaired components; the preserved pre-repair components satisfy five and fail ten (Table~\ref{tab:diagnostics}). The five clean controls conform under both implementations. The ten fault/edge probes were selected to expose known weaknesses, so the comparison must not be reported as an accuracy improvement from 33\% to 100\%. The diagnostic rows reproduce exactly on repeated execution after excluding runtime metadata. Preserved snapshots can be replayed with hash checking without requiring Git.

\begin{table}[!htb]
\centering
\caption{Conformance on author-selected component diagnostics. Counts are not population accuracy.}
\label{tab:diagnostics}
\begin{tabular}{lrrr}
\toprule Component & Cases & Pre-repair conforming & Repaired conforming\\ \midrule
Numeric factory & 2 & 1 & 2\\
Promotion boundary & 7 & 3 & 7\\
Overlap audit & 3 & 0 & 3\\
Near deduplication & 3 & 1 & 3\\ \midrule
Total & 15 & 5 & 15\\ \bottomrule
\end{tabular}
\end{table}

The refreshed overlap audit checks all 100,000 local training records against 1,280 evaluation cases across nine releases and detects no identity/source/exact-normalized collisions under its implemented comparisons. A separate integrity check verifies all nine evaluation manifests and 27 listed artifacts against the index and release metadata, with no discrepancies. Shared generators, paraphrases and templates remain outside this result; different numeric seeds do not establish an independent generalization benchmark.

A current-content quarantine scan over the 100K local records returns no implemented-pattern or retained-flag findings and no scan/schema errors. This does not estimate privacy-detector recall. Ordinary and supplemental Gold promotion are both covered by current checks; the low-level release writer remains a trusted construction API and is not itself a complete approval authority. Regression tests additionally cover metadata dispatch changes, invalid legacy numeric inputs, and promotion of old Silver releases. These software tests support the scoped behavior without supplying independent human validation.

\subsection{Incremental checks on constructed faults}
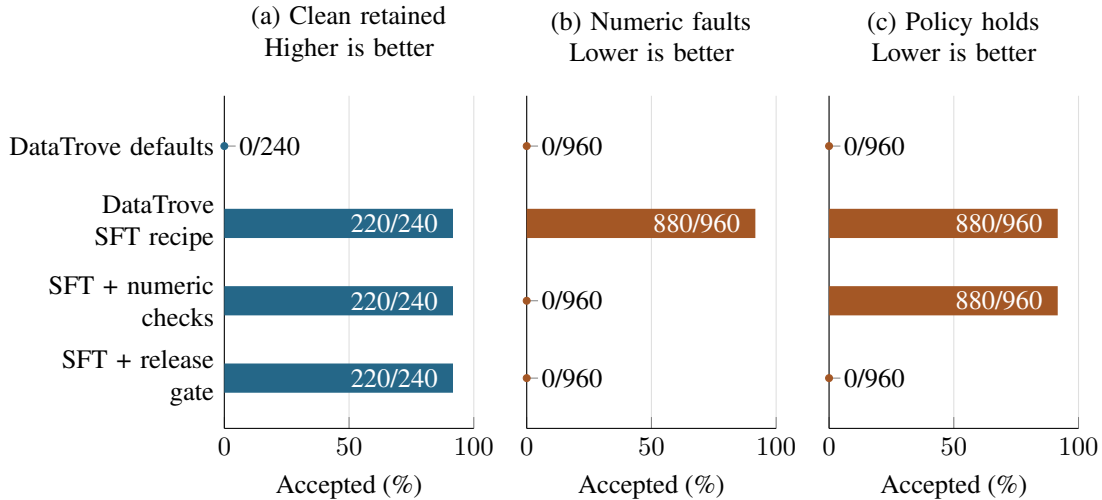
\begin{figure}[!htb]
\centering
\begin{lrbox}{\forgefigurebox}
\begin{tikzpicture}
\begin{groupplot}[group style={group size=3 by 1,horizontal sep=.70cm},width=3.30cm,height=4.4cm,scale only axis,xmin=0,xmax=100,ymin=-.65,ymax=3.65,xtick={0,50,100},ytick={0,1,2,3},xlabel={Accepted (\%)},tick label style={font=\small},label style={font=\small},title style={font=\small,align=center,text width=3.30cm,yshift=3pt},axis x line*=bottom,axis y line*=left,xmajorgrids=true,grid style={gray!25},clip=false]
\nextgroupplot[title={(a) Clean retained\\{\small Higher is better}},yticklabels={{SFT + release\\gate},{SFT + numeric\\checks},{DataTrove\\SFT recipe},{DataTrove defaults}},yticklabel style={text width=2.9cm,align=right}]
\addplot[xbar,bar width=11pt,draw=none,fill=forgeblue] coordinates {(0.0000000000,3) (91.6666666667,2) (91.6666666667,1) (91.6666666667,0)};
\addplot[only marks,mark=*,mark size=1.3pt,color=forgeblue] coordinates {(0,3)};
\node[anchor=west,font=\small] at (axis cs:2,3) {0/240};
\node[anchor=east,font=\small,text=white] at (axis cs:89.6666666667,2) {220/240};
\node[anchor=east,font=\small,text=white] at (axis cs:89.6666666667,1) {220/240};
\node[anchor=east,font=\small,text=white] at (axis cs:89.6666666667,0) {220/240};
\nextgroupplot[title={(b) Numeric faults\\{\small Lower is better}},yticklabels={},]
\addplot[xbar,bar width=11pt,draw=none,fill=forgeorange] coordinates {(0.0000000000,3) (91.6666666667,2) (0.0000000000,1) (0.0000000000,0)};
\addplot[only marks,mark=*,mark size=1.3pt,color=forgeorange] coordinates {(0,3)};
\node[anchor=west,font=\small] at (axis cs:2,3) {0/960};
\node[anchor=east,font=\small,text=white] at (axis cs:89.6666666667,2) {880/960};
\addplot[only marks,mark=*,mark size=1.3pt,color=forgeorange] coordinates {(0,1)};
\node[anchor=west,font=\small] at (axis cs:2,1) {0/960};
\addplot[only marks,mark=*,mark size=1.3pt,color=forgeorange] coordinates {(0,0)};
\node[anchor=west,font=\small] at (axis cs:2,0) {0/960};
\nextgroupplot[title={(c) Policy holds\\{\small Lower is better}},yticklabels={},]
\addplot[xbar,bar width=11pt,draw=none,fill=forgeorange] coordinates {(0.0000000000,3) (91.6666666667,2) (91.6666666667,1) (0.0000000000,0)};
\addplot[only marks,mark=*,mark size=1.3pt,color=forgeorange] coordinates {(0,3)};
\node[anchor=west,font=\small] at (axis cs:2,3) {0/960};
\node[anchor=east,font=\small,text=white] at (axis cs:89.6666666667,2) {880/960};
\node[anchor=east,font=\small,text=white] at (axis cs:89.6666666667,1) {880/960};
\addplot[only marks,mark=*,mark size=1.3pt,color=forgeorange] coordinates {(0,0)};
\node[anchor=west,font=\small] at (axis cs:2,0) {0/960};
\end{groupplot}
\end{tikzpicture}
\end{lrbox}
\fitforgefigure
\caption{Retention and acceptance on the 2,400 constructed test rows. Labels give exact accepted/total counts; all recipes accept 0/240 repetition probes. Clean retention is desirable, whereas accepting the injected numeric faults or release-policy holds is undesirable. Treatments are paired within 240 scenarios. Policy holds are eligibility conditions, not necessarily financial errors; four treatments alter metadata unavailable to the text-only baseline. The final arm is a composed record gate, not the full pipeline.}
\label{fig:curation}
\end{figure}

Figure~\ref{fig:curation} shows category acceptance with exact counts for the 2,400 test rows. The default web-text filters reject every clean control, illustrating an unsuitable operating point for short numerical instruction examples. The prespecified SFT recipe retains 220 of 240 controls and accepts 880 of 960 numeric-inconsistency probes. Adding the numeric validator rejects all those numeric probes but leaves 880 policy-held rows accepted. Adding the release-record gate rejects all constructed numeric faults and policy holds. Every arm rejects the 240 repetition probes.

For the native contracts, visible prompt and answer mutations are rejected by comparison with a regenerated canonical representation. This demonstrates contract enforcement, not semantic validation of independently authored questions or valid paraphrases. Separate arithmetic recomputation checks structured values under the shared formula specification. Likewise, the metadata-only treatments give the composed gate information absent from the text baseline; the figure reports incremental mechanisms, not equal-information classifier superiority.

The twenty rejected clean controls are \emph{every} cash-conversion-cycle control, while all twenty controls in each of the other eleven families survive (Table~\ref{tab:retention}). Repeated derivation/substitution content triggers the repetition filter's duplicate-line character criterion for ten text-format controls and its duplicated five-gram criterion for ten JSON-format controls. Both augmented arms inherit these losses. The aggregate 20/240 (8.3\%) clean rejection therefore conceals complete loss of one tested task family. No setting is changed after observing this result. Retention should be evaluated as coverage of useful tasks and representations, as well as an overall percentage.

\begin{table}[!htb]
\centering\small
\caption{Clean-control retention by family and representation. The SFT recipe and both augmented arms have identical counts. These are the existing 240 controls, not an additional experiment.}
\label{tab:retention}
\begin{tabular}{lrrr}
\toprule Formula group & Text retained & JSON retained & Total retained\\ \midrule
Cash conversion cycle & 0/10 & 0/10 & 0/20\\
Each of the other eleven families & 10/10 & 10/10 & 20/20\\ \midrule
All twelve families & 110/120 & 110/120 & 220/240\\ \bottomrule
\end{tabular}
\end{table}

There are no exact numeric-parameter overlaps between the 240 test scenarios and either the 24 development scenarios or the 15,000 existing Gold cases. This is parameter separation, not template-independent evaluation. Repeated decision hashes agree across three passes. Median warm pass times over all 2,640 development-plus-test rows are 3.47\,s for the SFT recipe, 3.83\,s with numeric checks, and 3.89\,s with release-record checks. These fixed-order, single-process observations do not estimate distributed throughput or broad comparative efficiency.

\subsection{Release boundary: rejection and acceptance limits}
The three ordinary controls produce Gold and eleven selected negative cases are blocked (Table~\ref{tab:release}). All four constrained semantic variants are rejected by the canonical numeric contract. These are losses of valid representations under the constructed equivalence criterion, even though rejection conforms to the implemented producer contract. They are not a population false-rejection rate. The JSON formatting control passes because equality is assessed after parsing.

\begin{table}[!htb]
\centering\small
\caption{Nineteen dependent persisted-release cases. Counts are case tallies; all semantic variants target the DSCR record.}
\label{tab:release}
\begin{tabularx}{\textwidth}{Xrr}
\toprule Case category & Accepted & Rejected\\ \midrule
Complete source/record approvals; JSON formatting & 3 & 0\\
Partial, overlapping, expired, wrong-parent approvals & 0 & 4\\
JSONL/Parquet byte tampering & 0 & 2\\
Wrong answer with stale flag and fresh approval & 0 & 1\\
Current or stored sensitive findings & 0 & 2\\
Missing or changed evidence before binding & 0 & 2\\
Equivalent wording, unit spelling, derivation & 0 & 4\\
Evidence modified after binding & 1 & 0\\ \midrule
Total & 4 & 15\\ \bottomrule
\end{tabularx}
\end{table}

Changing external evidence after binding still permits ordinary Gold promotion: the unchanged bound registry is trusted. This is a freshness limitation when deployment policy requires the current external evidence. Under snapshot approval, the relevant requirement is preservation and protection of the approved evidence version; the experiment does not validate an operational snapshot store or revocation service. All four accepted releases satisfy the artifact/content/lineage and immutable-destination checks. Both runs match in result bytes and all 357 files in each hash inventory, excluding the inventory itself. Nine runner tests pass. The experiment excludes low-level builder misuse, concurrent writers, filesystem access controls, and supplemental Gold, which has a different evidence-read boundary.

\subsection{Processing cost and controlled recovery}
All 27 planned uninterrupted stage runs complete with stable source hashes and no omissions. Figure~\ref{fig:performance} shows processing time and memory across scales; Table~\ref{tab:performance} in Appendix~\ref{app:measurements} preserves the numerical measurements and output sizes. At 100K raw rows, median stage times are 31.36\,s for audit, 13.48\,s for Bronze, and 4.02\,s for qualification. Their observed ranges are 31.35--33.60, 13.42--14.03, and 3.68--4.05\,s respectively. These stages write approximately 197.9, 242.7, and 45.8\,MiB of logical output at that scale, versus an 87.0\,MiB raw source; the output sizes describe retained artifact footprint, not bytes written by the storage device.

\begin{figure}[!htb]
\centering
\begin{lrbox}{\forgefigurebox}
\begin{tikzpicture}
\begin{groupplot}[group style={group size=2 by 1,horizontal sep=1.85cm},width=5.40cm,height=4.4cm,scale only axis,xmode=log,log basis x=10,xmin=750,xmax=135000,xtick={1000,10000,100000},xticklabels={1K,10K,100K},xlabel={Raw input rows, $N$ (log scale)},tick label style={font=\small},label style={font=\small},title style={font=\small,align=center,text width=5.40cm,yshift=3pt},grid=major,grid style={gray!25},minor tick num=0,axis x line*=bottom,axis y line*=left]
\nextgroupplot[title={(a) Processing time\\{\small Median and observed min--max}},ymode=log,ymin=.02,ymax=60,ytick={.03,.1,.3,1,3,10,30},yticklabels={0.03,0.1,0.3,1,3,10,30},ylabel={Stage time (s, log scale)},legend to name=forgecostlegend,legend columns=3,legend style={draw=none,font=\small},legend cell align=left]
\addplot[color=forgeblue,mark=*,solid,line width=.8pt,mark size=2pt,error bars/.cd,y dir=both,y explicit] coordinates {
(1000,0.224772749934) += (0,0.00436225009616) -= (0,0.00147816690151)
(10000,2.40326050005) += (0,0.00861524988431) -= (0,0.00312929204665)
(100000,31.3556012081) += (0,2.24319620896) -= (0,0.00556645798497)
};
\addlegendentry{Audit}
\addplot[color=forgeorange,mark=square*,dashed,line width=.8pt,mark size=2pt,error bars/.cd,y dir=both,y explicit] coordinates {
(1000,0.124271083972) += (0,0.0196301660035) -= (0,0.000238626031205)
(10000,1.23498520802) += (0,0.00826812605374) -= (0,0.00882600003388)
(100000,13.484673042) += (0,0.541742167086) -= (0,0.0662431669189)
};
\addlegendentry{Bronze migration}
\addplot[color=forgepurple,mark=triangle*,dotted,line width=.8pt,mark size=2pt,error bars/.cd,y dir=both,y explicit] coordinates {
(1000,0.0379382090177) += (0,0.000192623934709) -= (0,0.000363333965652)
(10000,0.353113833) += (0,0.00459887599573) -= (0,0.00234058289789)
(100000,4.02160266601) += (0,0.0315165419597) -= (0,0.341372248949)
};
\addlegendentry{Qualification}
\nextgroupplot[title={(b) Peak memory\\{\small Maximum across three repetitions}},ymin=0,ymax=70,ytick={0,20,40,60},ylabel={Peak RSS (MiB)}]
\addplot[color=forgeblue,mark=*,solid,line width=.8pt,mark size=2pt] coordinates {(1000,44.109375) (10000,46.90625) (100000,50.40625)};
\addplot[color=forgeorange,mark=square*,dashed,line width=.8pt,mark size=2pt] coordinates {(1000,48.8125) (10000,53.4375) (100000,56.609375)};
\addplot[color=forgepurple,mark=triangle*,dotted,line width=.8pt,mark size=2pt] coordinates {(1000,41) (10000,42.171875) (100000,44.75)};
\end{groupplot}
\end{tikzpicture}
\end{lrbox}
\fitforgefigure
\par\vspace{6pt}\ref{forgecostlegend}\par
\caption{Local processing measurements on the authored workload, with three uninterrupted repetitions per stage and size. Left: median stage time and observed minimum--maximum bars (not confidence intervals); small ranges can be hidden by markers. Right: the largest observed peak RSS in each group of three runs, on a linear scale starting at zero. Time excludes interpreter startup/imports; peak RSS includes them. Qualification reads $0.9N$ rows after deduplication. Connecting lines guide the eye; they do not fit or extrapolate a scaling law.}
\label{fig:performance}
\end{figure}
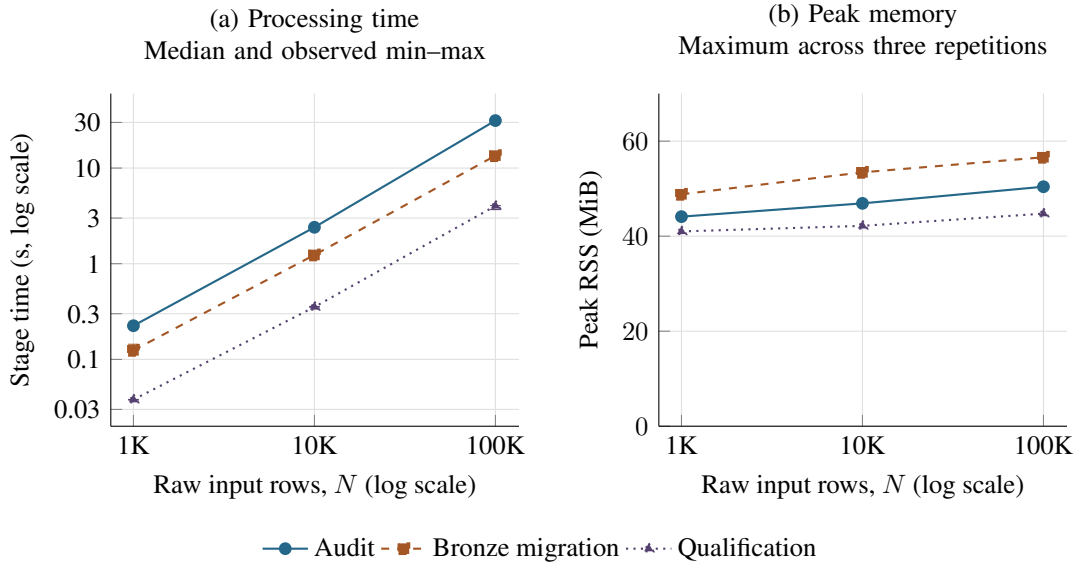

Exactly 10\% of raw rows are recognized as duplicates; migration produces the expected 90\% typed rows without migration errors. The raw workload contains no initially canonical, schema-valid records, and all migrated rows retain a \texttt{QUARANTINE} route. Qualification records missing-temporal-metadata warnings for the CPT rows and retains review-required rights status. These measurements therefore cover staging and rule execution, not production of approved training data. All three 10K interruption/resume comparisons match their uninterrupted substantive rows and summaries. The result supports the tested checkpoint boundaries; it says nothing about uncommitted-write durability, near deduplication, tokenizer cost, approval adjudication, or full release-builder memory.

\subsection{Reproducibility and scope}
The installed code-only distribution passes all 182 shipped tests in the recorded fresh-environment check. A separate data-inclusive extraction reproduces selected audit, overlap, and release results on the same host. The suites overlap and do not constitute independent replication; Appendix~\ref{app:reproduction} gives their exact scope. The full research inputs remain private, and the code package alone cannot reconstruct all reported results.

An exploratory review of 200 retained records by two same-model AI agents flags possible usefulness and evidence problems despite arithmetic agreement. These unweighted judgments motivate qualified review; they are not expert labels or population defect estimates (Appendix~\ref{app:ai}).

\section{Limitations and claims boundary}
The verified dataset is synthetic and comes from a single project. Numerical agreement across two implementations does not detect a shared specification error. Most nonnumeric domains have only operational metadata and domain-specific rules, not independent expert truth labels. Pattern-based sensitive-data scanning has limited coverage, and lexical deduplication cannot establish semantic uniqueness. The stricter numeric contract rejects all four selected equivalent representations; this small DSCR-based probe cannot estimate the frequency of such loss in other tasks or natural data.

The overlap audit does not establish separation of templates, source knowledge, or model pretraining data. Hashes establish identity relative to supplied inputs, not source correctness or authenticated approval. Study designs were informed by inspected code and observed defects; the later frozen experiments remain internal and may underrepresent unknown failure modes. The nineteen release cases share two source records and do not test a hostile deployment or authenticated approval service. The comparative study uses two actual DataTrove blocks, not a tuned competitive corpus pipeline; its shared generators and native contracts can favor Forge's validators. Its rejection labels derive from explicit perturbations and policies, not blinded expert truth. Likewise, two same-model AI reviewers can agree and still be wrong.

The full-corpus bundles, independent human labels, natural-defect curation comparison, and matched-budget model experiments remain unavailable or unexecuted. Performance observations are limited to one laptop and a repetitive authored workload; a fresh virtual environment still shares the host and package cache. The full research artifacts remain private, limiting external verification until confidential reviewer access is arranged. Appendix~\ref{app:future} records the additional evaluations needed for stronger claims. These limits permit a bounded systems claim, not broad financial validity, framework superiority, or causal downstream model benefit.

\section{Broader impact and assistance disclosure}
Binding validation evidence to released content can make unnoticed changes and unresolved release conditions easier to audit. However, financial training data can still contain unsupported decisions, omitted risks, misleading precision, or jurisdictional assumptions even when its arithmetic is correct. Using a clean gate report as a substitute for substantive financial review could propagate those weaknesses to consequential applications. Strict contracts and generic filters can also remove useful examples and narrow coverage. Our reports therefore separate arithmetic, policy, uncertainty, and retention outcomes; no artifact certifies suitability for lending, compliance, or customer decisions.

AI assistants were used to inspect and modify software, construct experimental runners, assist manuscript drafting, and conduct the explicitly labeled exploratory review. Separate reviewer agents shared the same underlying model family and received fresh review contexts. They are not authors or qualified human participants. Their labels are retained as model outputs, and responsibility for the final scientific claims, authorship, release permissions, and submission rests with the human authors.

\section*{Funding and competing interests}
This work was funded by Arya.ai. Both authors are affiliated with Arya.ai, the developer of the system studied. No external financial experts participated in the reported AI-assisted review.

\section{Conclusion}
This case study connects financial instruction-record checks to manifest-scoped release approval and measures their interaction. Fresh native-contract validation rejects the selected stale-content faults, while the composed filtering recipe removes every tested clean cash-conversion-cycle example and the strict contract rejects four equivalent DSCR representations. The ordinary and supplemental promotion paths also differ in when they read approval evidence. These findings support explicit interface specifications, protected approval inputs, and retention reporting by task family. Integrity audits and controlled recovery provide supporting implementation evidence. Independent examples, qualified labels, and equal-input mechanism comparisons are the next tests of generality; model benefit requires a separate training study.

\section*{Data and code availability}
A standalone code distribution, version 0.1.0, is prepared under the Apache License 2.0, copyright Arya.ai, as \artifact{arya-forge-0.1.0-source-20261025.tar.gz}. It contains the allowlisted implementation, documentation, integrity manifest, and portable tests, with a separately built wheel. Publication at a public repository has not yet occurred. The synthetic populations, case-level results, AI annotations, approval fixtures, and research evidence remain private and are not covered by the code distribution's license. A separate confidential reproduction package preserves these supporting materials. Controlled reviewer access arrangements have not been established. No public dataset, code-hosting URL, or artifact DOI is claimed.

\appendix
\section{Exploratory AI-assisted assessment}\label{app:ai}
\subsection{Exploratory AI-assisted record review}
At the project owner's request, two fresh, same-base-model assistant agents separately review the 200-record retained packet. Each receives only the common rubric and visible conversations, without the coordinator's domain mapping, pipeline scores, factory implementations, reference judgments, the manuscript, or the other agent's annotations. Each may use independently written arithmetic code; no external factual source is authenticated by this procedure. Labels are locked separately before coordinator analysis, and the original blank human-review packet is preserved.

The rubric covers numeric correctness, support by supplied evidence, jurisdiction/date applicability, and training usefulness, with explicit uncertainty and not-applicable categories. Analysis reports paired categorical agreement and Cohen's $\kappa$, including those categories, and retains disagreements without manufacturing a consensus truth label. These are AI judgments from a shared model family, not independent financial experts, a human-subject study, or established data-quality error rates. The model/API revision and sampling seed were not pinned, so preserving the labels enables analysis replay rather than deterministic recreation of the model judgments. Repeated templates and correlated model assumptions further limit their evidential strength.

\subsection{AI review agreement and unresolved findings}
Both agents complete and lock all 200 records. Reviewer A labels 151 useful, 34 not useful, and 15 uncertain; reviewer B labels 128 useful, 32 not useful, and 40 uncertain. They jointly label 31 records not useful. At least one adverse label appears on 61 records, and at least one rubric dimension differs on 58 records. These are unweighted descriptions of the equally allocated sample, not estimated defect rates in the 100K population.

\begin{table}[!htb]
\centering
\caption{Descriptive agreement of two same-model AI reviewers over 200 paired records. Uncertain and not-applicable labels remain categories; agreement is not accuracy.}
\label{tab:ai-review}
\begin{tabular}{lrr}
\toprule Dimension & Raw agreement & Cohen's $\kappa$\\ \midrule
Numeric correctness & 80.0\% & 0.573\\
Evidence support & 84.0\% & 0.698\\
Jurisdiction/date applicability & 91.0\% & 0.832\\
Training usefulness & 86.5\% & 0.716\\ \bottomrule
\end{tabular}
\end{table}

Reviewer A records 395 recomputed or checked values, including 392 explicit numerical comparisons; reviewer B records 395 explicit comparisons. Neither checker reports a mismatch, conditional on its formulas and assumptions. The counts differ because three entries in A's filing fallback record supplied source values without comparing them with a candidate output; the final audit preserves this distinction rather than treating every computed entry as a comparison. Nevertheless, their rubric interpretations differ: reviewer B marks forty credit confidence scores uncertain because their numerical meaning is unspecified, while reviewer A treats them as separate uncalibrated signals and assesses the displayed financial calculations. This accounts for all forty numeric-label disagreements. It illustrates measurement ambiguity rather than forty discovered arithmetic errors.

Their notes flag omitted credit risks or covenant breaches, AML closure decisions without a supplied closure rule, benign requests answered only with an offer of assistance, unstated interest conventions, and a materiality ranking based on unavailable weights. These are candidates for qualified review, not adjudicated findings. The analysis preserves both original labels and does not convert agreement into a reference answer. Automated arithmetic conformance therefore coexists with unresolved judgments about completeness, evidence, and usefulness; the study does not collapse them into a single quality score.

\section{Further evaluation protocol}\label{app:future}
These extensions are unexecuted and are not prerequisites for interpreting the bounded systems observations as reported. They are required before stronger claims about expert-validated quality, generalization, or downstream utility. The original prospective plan is retained separately from dated execution updates; no experiment is described as externally preregistered.

\paragraph{Independent representation holdout.} Separate new cases from the retrospective defect probes and previously inspected controls. An author or panel independent of validator implementation should establish expected behavior for valid paraphrases, unit aliases, equivalent derivations, and faults in sign, scale, period, source selection, or assumptions. Cover every supported formula and more than one independently authored template family. Freeze development and holdout membership before tuning; changes prompted by the current cash-conversion-cycle and DSCR failures belong in development. Report unsupported formats separately, and choose sample sizes against a stated precision target. A new internally generated template set alone would not establish independent financial validation.

\paragraph{Equal-input mechanism comparison.} For content checking, give all arms the same structured record and compare schema/cached flags, arithmetic recomputation, canonical projection checking, and their composition. Preserve arithmetic-only and text-only ablations to identify what each rejects. For release control, hold content validators and approval inputs fixed while varying cached versus fresh checks and unbound versus manifest-bound coverage. Include held-out nonnumeric stale-grounding cases, approval lifecycle events, and both promotion entry points. Measure invalid acceptance, useful-content rejection, unsupported cases, and cost by scenario and family. Paired treatments and shared templates require clustered uncertainty rather than independent-row binomial intervals. These proposed arms have not been executed.

\paragraph{Qualified assessment.} A private collection package is prepared for the unchanged 200-record retained sample, with twenty records per domain, deterministic selection, and randomized presentation. Reviewers have not been selected and no human ratings have been collected. The plan calls for two qualified, independent human judgments per record, allowing domain-specialist panels, followed by separate adjudication. A twenty-record pilot (two per domain) precedes the remaining 180; generic rubric ambiguities may be resolved without disclosing counterpart case labels. Any changed rubric requires preserved earlier versions and independent pilot rerating. The four dimensions separate financial calculations, evidence support, jurisdiction/date applicability, and training usefulness. Forms begin blank, exclude AI annotations and private provenance, and allow reassignment outside competence. Analysis requires locked complete pairs; report per-domain counts, descriptive agreement, and explicitly weighted retained-mixture fractions without treating uncertain labels as errors. A separate excluded-record arm is needed for rejection error, with source/domain/reason strata frozen after the primary excluded corpus is available. This preparation is not a completed human study or an external preregistration.

\paragraph{Curation and model utility.} A broader comparison should hold source inventory, tokenizer, and training budget fixed across generic curation, added financial checks, and complete selection/release processing. Evaluate naturally occurring retained defects and useful-content losses using independent adjudication. Approval eligibility must not be equated with model accuracy. For model comparisons, freeze revision, tokenizer, optimizer, checkpoint-selection rule, total tokens, and three training seeds before execution; report a matched-record sensitivity analysis separately. Evaluate independently authored cases with source, organization, period, and template-family separation where applicable. External tasks such as FinQA require a compatible adapter and documented permitted use and overlap checks. Report paired results with uncertainty clustered by source or scenario family, retaining delivery failures and excluding selection exposure from the final holdout.

\paragraph{Systems and replication.} Extend heterogeneous document lengths and hardware beyond the current laptop workload. Measure near deduplication, tokenization, selection, and full release construction separately. Inject process termination between committed checkpoints and test chained recovery, concurrent use, and externally managed approval/revocation updates. The present post-checkpoint experiment does not cover those conditions. Independent replication requires access to the private study materials; a new environment on the original host is insufficient.

\section{Detailed processing measurements}\label{app:measurements}
\begin{table}[!htb]
\centering\small
\caption{Three repetitions per cell on the authored laptop workload. $N$ is raw input scale; qualification reads $0.9N$ canonical rows. Time excludes interpreter startup, while peak RSS includes it.}
\label{tab:performance}
\begin{tabular}{llrrr}
\toprule $N$ & Stage & Median seconds & Maximum RSS (MiB) & Output (MiB)\\ \midrule
1,000 & Audit & 0.225 & 44.1 & 2.0\\
 & Bronze & 0.124 & 48.8 & 2.5\\
 & Qualification & 0.038 & 41.0 & 0.5\\
10,000 & Audit & 2.403 & 46.9 & 19.7\\
 & Bronze & 1.235 & 53.4 & 24.3\\
 & Qualification & 0.353 & 42.2 & 4.6\\
100,000 & Audit & 31.356 & 50.4 & 197.9\\
 & Bronze & 13.485 & 56.6 & 242.7\\
 & Qualification & 4.022 & 44.8 & 45.8\\ \bottomrule
\end{tabular}
\end{table}

\section{Reproduction scope and missing evidence}\label{app:reproduction}
\subsection{Recorded reproduction checks}
The fifteen component diagnostics reproduce from an extracted, hash-checked source archive after installing pinned dependencies into a new virtual environment. Of the 190 focused tests that pass in the original workspace, 169 also pass without private dataset fixtures. The initial source-only run records 21 failures caused by unavailable release fixtures; these are explicitly enumerated and excluded from the portable rerun rather than counted as passes. Both runs and installation logs are retained. This verifies dependency installation and portable checks on the same host, not an independent laboratory replication. The earlier broader caller-suite run remains incomplete.

A refreshed code-only distribution contains 167 files, including implementation, configuration, documentation, portable tests, and its integrity manifest. It excludes research populations, review labels, governance evidence, and Git history; its only JSONL fixture contains six explicitly synthetic regression probes. A non-editably installed wheel built from a clean archive extraction passes all 182 shipped tests with zero skips or failures in a fresh environment outside the checkout, including generation of twelve Bronze examples and release reporting. Packaging regressions verify that the installed CLI loads the bundled audit configuration and that generated-data ignore rules preserve evaluation source files. The 113 included runtime source identities match the evaluated snapshot; two evaluation package initializers retain the narrowed public interfaces. Two adapters requiring the separate serving/orchestration application are outside the standalone toolkit. These packaging checks overlap with the repository tests and must not be summed as unique coverage. The refresh ran on Python 3.13.7 and macOS 27.0.1 on ARM64; Linux CI and optional tokenizer integration were not executed.

In the recorded private reproduction check, a confidential data-inclusive archive is extracted into a separate local directory. With the existing host environment, its portable wrapper reproduces the per-release and per-formula 100K audit outputs and the 1,280-case overlap result. All 190 focused tests pass from that data-inclusive extraction with zero exclusions, resolving the source-only fixture omissions. The extracted release-boundary result bytes, locked-label analysis, and component diagnostics agree with their recorded counterparts; one 1,000-row audit worker is a performance smoke check. This verifies selected relocation paths and supplied data, not a new-machine replication, a complete benchmark rerun, or recreation of AI judgments. The subsequent manuscript and public-package refresh preserves the empirical inputs; it is not counted as another execution of those 190 tests.

\subsection{Unavailable corpus evidence}
The primary full-corpus audit, qualification, candidate, and promotion bundles were not located in the local evidence inventory. Larger counts described in repository documentation are therefore excluded from this paper's verified totals. The documented exclusion process combines duplication, validation, authorization policy, and mixture selection; its aggregate rejection cannot be interpreted as an invalid-data percentage. In particular, exclusion for unbound rights evidence does not establish that a source legally forbids use.

\section{Research artifact documentation}\label{app:data}
\paragraph{Purpose and composition.} The study materials support checks of representation consistency, supported arithmetic contracts, release decisions, and local execution. The 100,000 retained records and domain counts are described in Table~\ref{tab:mixture}. Additional authored cases support the filter, release-boundary, and processing experiments; they do not form an independently authored financial benchmark. The material is not intended to certify suitability for lending, compliance, investment, or customer decisions.

\paragraph{Collection and labels.} Synthetic generators, source identifiers, transformation lineage, and validator versions are preserved in the source and release manifests. Synthetic source markers do not themselves prove ownership or factual validity. Historical validation metadata remain unchanged; the current audit supplies separate retrospective observations. The exploratory review labels are AI outputs. There are no collected qualified-human labels, and no human-participant approval or exemption is asserted for a future study.

\paragraph{Access and maintenance.} Research artifacts remain private at the project owner's direction. The public code license grants no release of the separate research materials. A controlled-access contact, terms, hosting, and retention arrangements have not been established. A portable core experiment would need cleared synthetic generators, protocols, runners, case-level outputs, and pinned dependencies; this is an outstanding dissemination step. Corrections should preserve earlier evidence and record new versions. The code-only distribution cannot reproduce the full empirical study without its private inputs.

\section{Artifact map}\label{app:artifacts}
Within the confidential archive, the empirical root is \artifact{outputs/forge-research-20261007/}. Its \artifact{REPRODUCE.txt} maps the audit, component diagnostics, filter, processing, and release-boundary runners to their frozen evidence. The \artifact{final-audit/} directory contains independent representation checks; \artifact{ai-review/} contains locked model labels, while \artifact{review/} and \artifact{human-review/} contain the separate review preparations. These are private evidence locations, not public download links.

The current submission source is \artifact{manuscript/paper.tex} in the delivery workspace; the arXiv source bundle contains it as \artifact{paper.tex}, with inline figures and bibliography. The later research revision changes presentation and analysis of existing results, without replacing the archived experiments or frozen implementation. Manuscripts inside the preserved confidential archive are earlier versions. Only the separately identified Apache-licensed code distribution is a public-release candidate; neither the surrounding research archive nor the coordinator's review kit is the public code release.

\end{document}